\documentclass{article}
\usepackage{spconf,amsmath,graphicx,booktabs,pmboxdraw,stmaryrd}

\title{SPELLING CORRECTION THROUGH REWRITING OF NON-AUTOREGRESSIVE ASR LATTICES}
\makeatletter
\def\@name{ \emph{Leonid Velikovich},  \emph{Christopher Li}, \emph{Diamantino Caseiro}, \emph{Shankar Kumar}, \\ \emph{Pat Rondon}, \emph{Kandarp Joshi}, \emph{Xavier Velez}}
\makeatother
\address{Google, Inc}

\begin{document}
%

\maketitle

\begin{abstract}
\noindent For end-to-end Automatic Speech Recognition (ASR) models, recognizing personal or rare phrases can be hard. A promising way to improve accuracy is through spelling correction (or rewriting) of the ASR lattice, where potentially misrecognized phrases are replaced with acoustically similar and contextually relevant alternatives.  However, rewriting is challenging for ASR models trained with connectionist temporal classification (CTC) due to noisy hypotheses produced by a non-autoregressive, context-independent beam search.

We present a finite-state transducer (FST) technique for rewriting wordpiece lattices generated by Transformer-based CTC models. Our algorithm performs grapheme-to-phoneme (G2P) conversion directly from wordpieces into phonemes, avoiding explicit word representations and exploiting the richness of the CTC lattice. Our approach requires no retraining or modification of the ASR model. We achieved up to a 15.2\% relative reduction in sentence error rate (SER)  on a test set with contextually relevant entities.
\end{abstract}

\section{INTRODUCTION}
\label{sec:intro}

ASR has transitioned towards a jointly trained end-to-end (E2E) architecture that bypasses the need for explicit phonetic representations~\cite{e2easr}. This has improved overall accuracy, but certain types of phrases pose a challenge:

\begin{itemize}
\item Homophonous phrases, where disambiguation requires context (``wall mount''/``Walmart'').
\item Phrases whose pronounciation is hard to infer from E2E training data (``call Mathijn'').
\item Foreign, novel, or rare phrases with unknown prior likelihood and pronunciation (``Kazi Mobin-Uddin'').
\end{itemize}

To address these challenges, ~\cite{Serrino2019, classlmwordmapping} proposed rewriting ASR lattices. These algorithms generate phonetically close and contextually relevant rewrites of the ASR hypotheses. Unlike first-pass solutions that require integration into the ASR beam search algorithm~\cite{apetar}, rewriting does not require modification of the ASR model.

Neural networks and large language models (LLMs) are increasingly used for spelling correction, but finite-state transducers offer a pragmatic alternative.  Like LLMs, they require no coupling with the ASR beam search, however they require no retraining, and can efficiently process the entire lattice as a graph, support custom pronunciations, and offer a level of explainability that simplifies debugging~\cite{fstcorrect}.

We extend the FST technique described in~\cite{Serrino2019}, which was applied to autoregressive lattices, to handle non-autoregressive CTC beam search, where tokens for any given frame are not conditioned on earlier tokens. Graphemic or wordpiece-based CTC beam search has the practical benefit of enabling the decoding of multiple frames independently~\cite{Prabhavalkar2017}; however, high lattice density requires new techniques for rewriting CTC lattices.
\section{PRIOR WORK}
\label{sec:prior}

Spelling correction of ASR hypotheses was shown to improve ASR quality \cite{sainath_19, Zhang2019}, particularly, the recall of named entities \cite{Le2020}. Contextual spelling correction uses additional information in the correction model, such as the device state, e.g., whether a voice assistant is currently playing music. \cite{Wang2021} proposed a sequence-to-sequence model that uses 
a context encoder to generate phrase embeddings that augment the text embeddings used by the attention in the decoder. \cite{Wang2022} employed both autoregressive and non-autoregressive models. 

Lattice augmentation techniques~\cite{Serrino2019,classlmwordmapping} operate on the whole word lattice, identifying spans where contextually relevant phrases are likely to occur, and match those to acoustically close candidates. To scale to the large number of hypotheses in a dense lattice, span identification and entity fuzzy matching use efficient Weighted FST (WFST) techniques. ~\cite{ponnusamy2020, mondegreen} present approaches that operate on $n$-best hypotheses and use domain knowledge derived from user interactions. ~\cite{mondegreen} uses neural candidate generation and replaces the entire transcript. LLMs have recently been proposed for spelling correction of ASR hypotheses. \cite{nvidia-antonova-2023} uses retrieval to find candidate corrections which are applied using a BERT~\cite{bert}-style neural model, while \cite{song2023} uses soft and hard prompts to reuse a general purpose LLM for the ASR spelling correction task. Non-postprocessing contextual ASR is usually tightly coupled with the ASR system, either during beam-search~\cite{apetar, Vasserman2016, Zhao2019ShallowFusionEC, facebook-chen2019a}, or in the ASR Model~\cite{pundak2018deep, facebook-chen2019b, Chang2021-zp}.

\section{ON NON-AUTOREGRESSIVE LATTICES}
\label{sec:lattice_description}

\label{sec:nonautoregressive}
For spelling correction of contextual ASR, we observed considerable differences between the lattices produced by autoregressive and non-autoregressive beam search. Non-autoregressive lattices contain a denser, noisier signal that is not well-handled by existing rewriting approaches. In this paper, we present techniques for rewriting non-autoregressive CTC lattices that not merely overcome these obstacles, but attain improved performance from the richer, more robust signal present in such lattices.

\begin{figure}[ht]
\centering
\includegraphics[width=\linewidth]{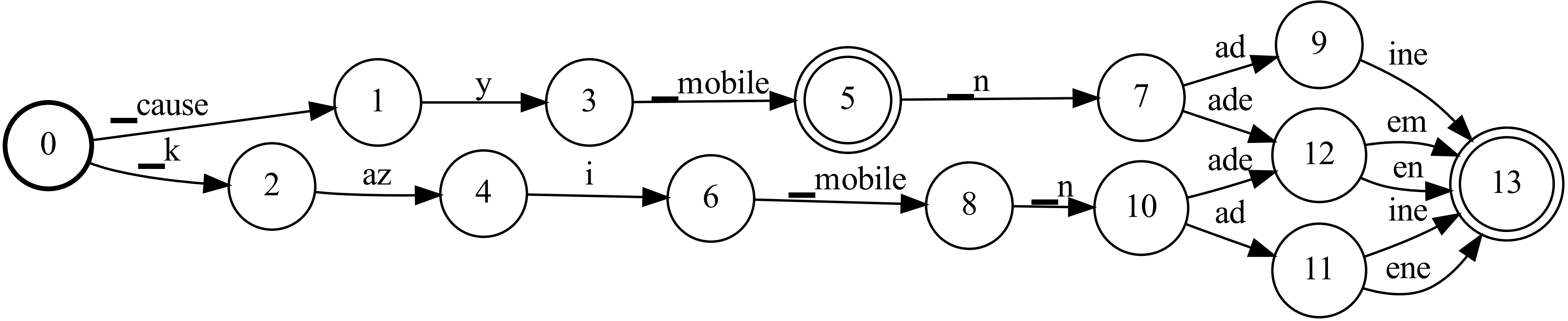}

\includegraphics[width=\linewidth]{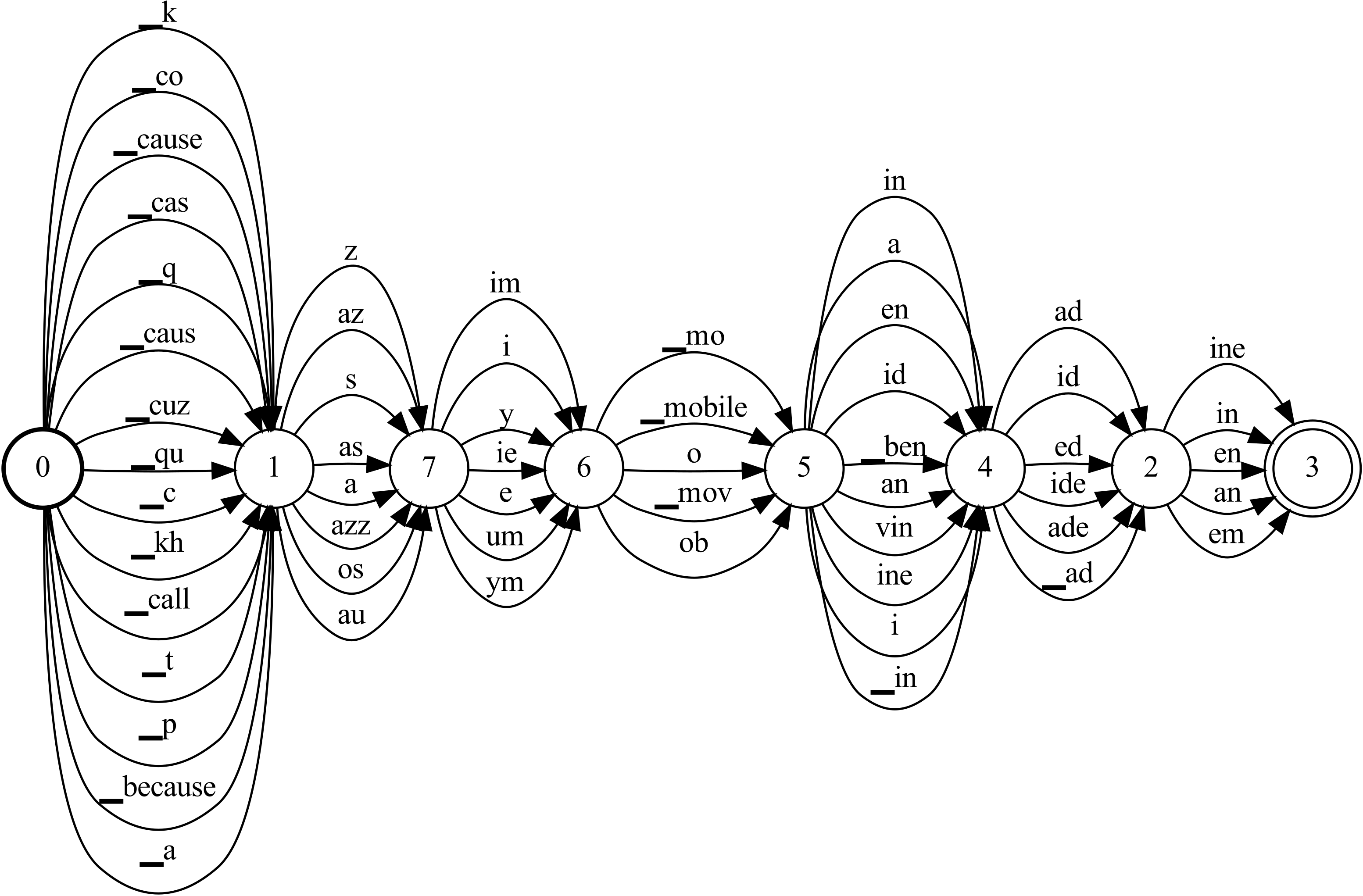}
\caption{Autoregressive (RNN-T; top) and non-autoregressive (CTC; bottom) lattices after incorrectly decoding a TTS utterance with imagined name "Kazi Mobin-Uddin". The lattices have 8 paths and 1.26 million paths, respectively.}
\label{fig:comparison_of_lattices}
\end{figure}

In non-autoregressive ASR, emitted tokens are conditionally independent from earlier or later tokens.  While this was shown to produce accurate 1-best results for CTC \cite{Prabhavalkar2017}), in the event of a misrecognition conditional independence presents challenges for lattice rewriting, as it results in a ``sausage lattice'' topology with exponentially many paths -- most of which are invalid lexically (wordpieces do not form valid words), semantically (words do not form coherent sentences), or even temporally (gaps or overlaps in time between adjacent tokens) -- see Fig.~\ref{fig:comparison_of_lattices}.  High path count and a prevalence of invalid paths make traditional word-based rewriting techniques impractical for a dense CTC lattice.

Nevertheless, we found an important advantage arising from the low influence of priors when using conditionally independent CTC decoding.  When an ASR misrecognition occurs, conditional independence prevents a ``domino effect'', where a mispredicted entity causes adjacent tokens to be wrong.  For example, if "Beth" in ``$call$ Beth'' is misrecognized as "bath", conditionality may also influence the adjacent word to be emitted as, e.g., ``$cold$ bath''. Avoiding this dependency makes rewriting using grammars such as ``$call$ \$CONTACT'' (\cite{Serrino2019}) more reliable.

Furthermore, conditional independence in CTC minimizes the harmful effects of label bias \cite{hannun2019label}, where mispredicted graphemes are acoustically skewed due to the influence of priors, causing overfitting to an incorrect token history.   Without label bias, even incorrect emitted graphemes capture the phonetics more faithfully, providing a better signal for rewriting compared with lattices produced by autoregressive ASR techniques such as classic hybrid ASR and RNN-T~\cite{rnnt}.

The above considerations make rewriting CTC lattices a promising area of research.
\section{REWRITING NON-AUTOREGRESSIVE LATTICES}
\label{sec:rewriting}

We first summarize the approach proposed by \cite{Serrino2019}, which we use as our baseline, to rewrite word lattices using FST operations. Then we describe techniques we developed for rewriting non-autoregressive CTC lattices.

\subsection{Word-based Lattice Rewriting}
\label{sec:identifyingspans}

\begin{figure}[ht]
\centering
\includegraphics[width=\linewidth]{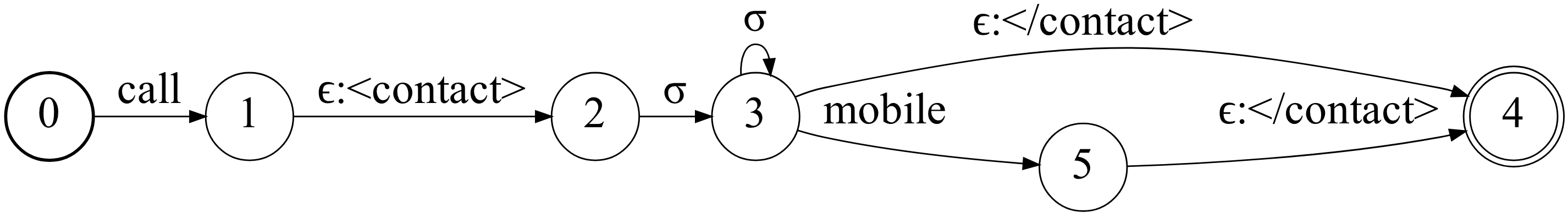}
\caption{A grammar-tagging FST for carrier phrases "call \$CONTACT" and "call \$CONTACT mobile".}
\label{fig:pattern_tagger}
\end{figure}

\begin{figure}[ht]
\centering
\includegraphics[width=150pt]{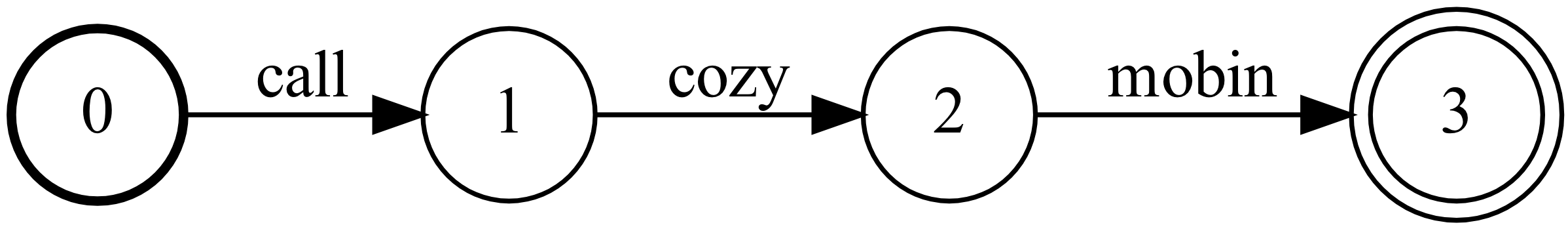}\\[10pt]
\includegraphics[width=\linewidth]{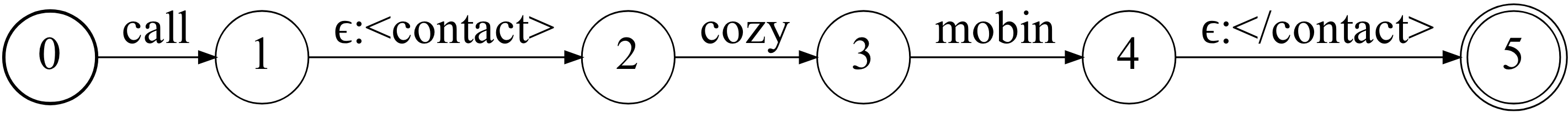}\\[6pt]
\includegraphics[width=110pt]{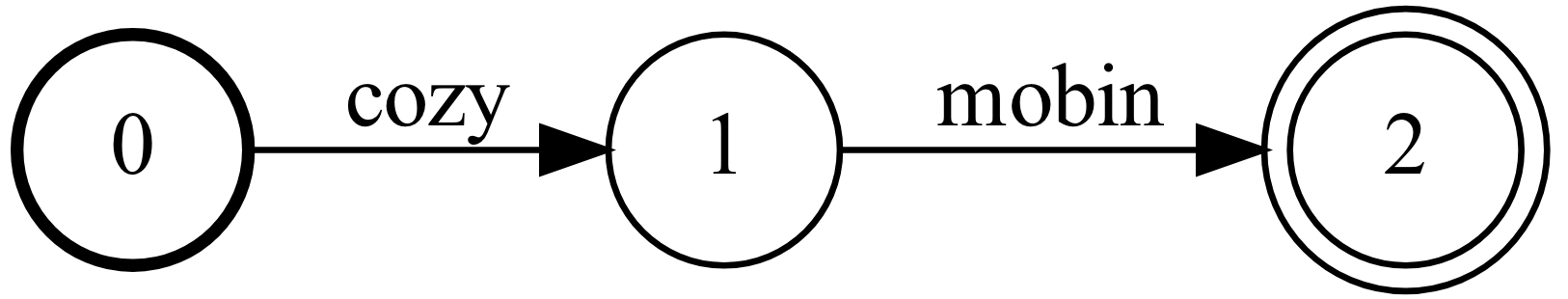}
\caption{A lattice with misrecognition ”call cozy mobin” (top); lattice after applying the grammar-tagging FST (middle) and extracting the tagged span for rewriting (bottom).}
\label{fig:cozy_tagged}
\end{figure}

Let $w$ denote a word lattice produced in the ASR first pass. We can identify potentially misrecognized phrases from $w$ using a grammar-tagging FST, $T$, via a special FST $\sigma$-composition ($\circ_\sigma$) that allows $\sigma$ in the right FST to match any token in the left FST \cite{mohri2002weighted}:
\begin{equation}
y \leftarrow w \circ_\sigma T    
\end{equation}

\noindent where $T$ identifies spans of one or more tokens in $w$ that are associated with entities denoted by nonterminals in predefined grammars. If $w$ contains "call cozy mobin" and $T$ includes ``call \$CONTACT'', then $y$ would contain "cozy mobin".  An example tagging FST is shown in Fig.~\ref{fig:pattern_tagger} and its application in Fig.~\ref{fig:cozy_tagged}.  Note that this step merely identifies a span to rewrite.

Let $G$ represent a G2P model that converts input graphemes to an FST consisting of phoneme representations in the X-SAMPA phonetic alphabet \cite{wells1995computer}. $G$ may be a transducer with a fixed vocabulary, or a model that can process any inputs.

\begin{figure}[ht]
\centering
\includegraphics[width=\linewidth]{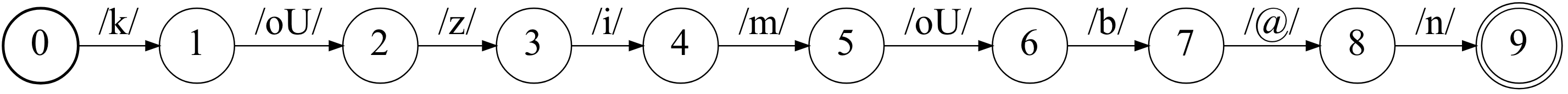}
\\[6pt]
\includegraphics[width=\linewidth]{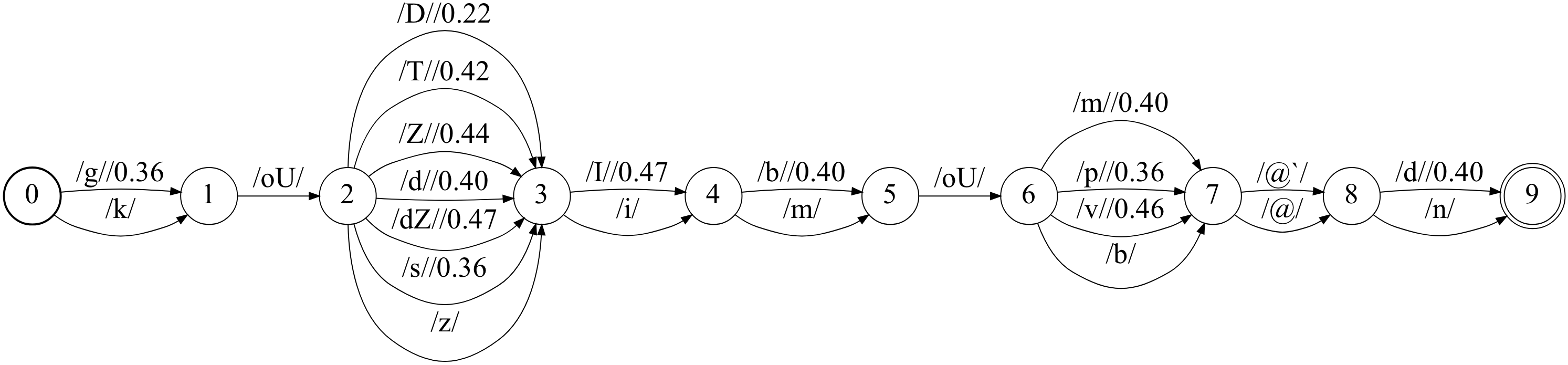}
\caption{Phoneme lattice before (top) and after (bottom) applying expansion with similar phonemes.}
\label{fig:cozy_phonemes_expanded}
\end{figure}

Let $P$ represent a phoneme-to-phoneme FST that enriches the lattice with acoustically close phonemes. Phoneme proximity is based on $L_2$ distances in a feature vector space whose features were designed by human linguists: place and manner of articulation, height, position and length (see Fig. ~\ref{fig:cozy_phonemes_expanded}).

\begin{figure}[ht]
\centering
\includegraphics[width=120pt]{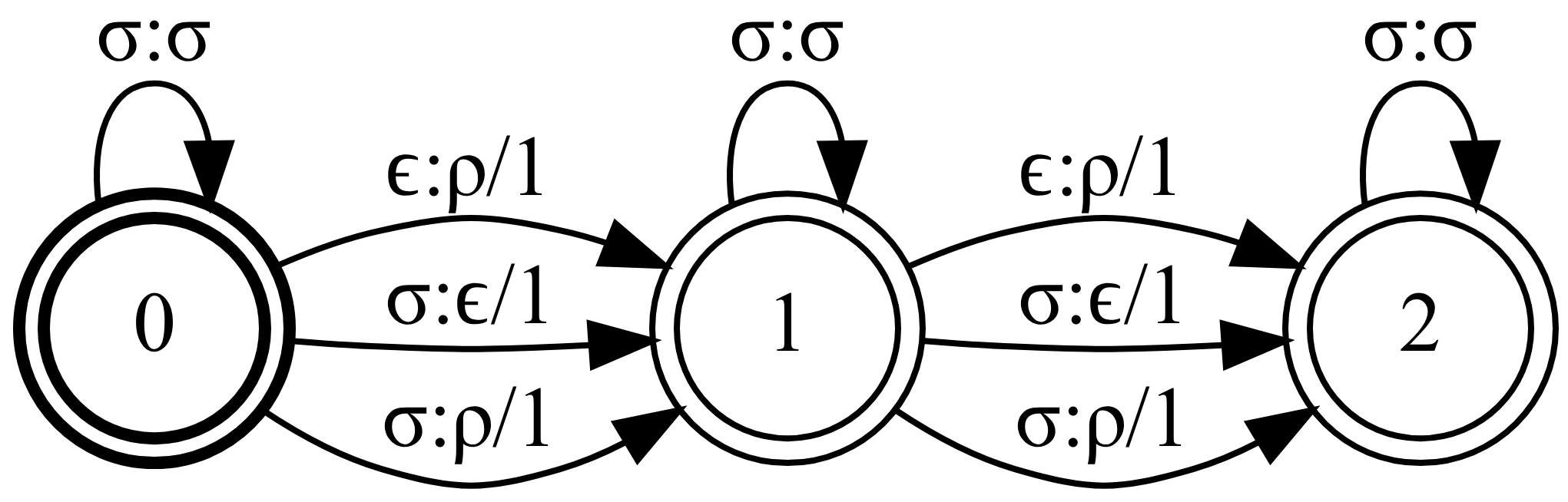}

\caption{An edit distance FST that allows at most 2 edit operations; each edit incurs a unit cost.}
\label{fig:edit_distance_2}
\end{figure}

Let $E$ represent a weighted edit-distance expansion FST that inserts additional arcs corresponding to a fixed number of phoneme edits (substitution, addition, replacement) of the input FST. The topology of a maximum $k$-edit FST is as follows: the states are $s_0 ... s_k$, where state $s_i$ represents $i$ cumulative edits performed so far. Each state has a self-arc with a $\sigma$ label representing a non-edit (exact match). Edits are represented by arcs transitioning from $s_i$ to $s_{i+1}$ with labels: $\epsilon:\rho$ (insertion), $\sigma:\epsilon$ (deletion), and $\sigma:\rho$ (substitution) as described in \cite{Serrino2019}. An example $2$-edit FST with a per-edit cost of 1 is shown in Fig.~\ref{fig:edit_distance_2}. For simplicity, we used a single fixed cost for all types of edits.

The resulting expanded phonemes FST is computed as:

\begin{equation}
    z \leftarrow \Downarrow(((y \circ G) \circ_\sigma P) \circ_\sigma E)
\end{equation}

\noindent where $\Downarrow$ projects the FST on output labels, $\circ$ denotes regular FST composition, and $\circ_\sigma$ denotes $\sigma$-composition. $z$ may contain special $\rho$ labels representing phoneme substitution.

Given a list of contextually-relevant entities, such as contact names, a phonemes-to-entity FST is constructed as follows. Let $s$ denote an FST containing the graphemes for each contextual phrase. We construct a phonemes-to-entity FST $s'$, where $inv$ denotes the FST inversion operation: 
\begin{equation}
s' \leftarrow inv(G) \circ s
\label{eqn:phonemes_to_entity_fst}
\end{equation}

Spelling correction is accomplished by performing the composition: 
\begin{equation}
h \leftarrow det(rmeps( \Downarrow(z \circ_\rho s')))
\label{eqn:recovery}
\end{equation}

\noindent where $det$ denotes FST determinization \cite{mohri2002weighted}, $rmeps$ denotes epsilon removal, $\circ_\rho$ denotes $\rho$-composition where $\rho$ labels in the left operand are consumed when no other labels match the right operand \cite{mohri2002weighted} (this enables using $\rho$ labels in $z$ for phoneme substitution).  $h$ contains the recovered contextually-relevant phrases. The candidate phrases are scored by adding the negative-log weights in the lattice path representing the candidate phrase.  The weights are a sum of first-pass CTC scores and any phonetic distance penalties applied during phoneme enrichment and edit distance steps.

In summary, the approach works by identifying spans in the lattice that may contain a misrecognized entity, extracting phonemes and performing entity retrieval over phonemes.


\subsection{Avoiding Word Representations}

A word-based FST spelling correction system requires converting a wordpiece lattice into words.  We found this approach to be suboptimal for non-autoregressive CTC wordpiece lattices, as it restricts our search space to a tiny subset of all paths in the CTC lattice.

For example, an utterance “call Kazi Mobin-Uddin” was misrecognized by a CTC model such that the 1-best path was “call co|z|y mo|ad|ine|an”, where | represents wordpiece boundaries. This path is lexically invalid and also acoustically distant from the truth. If we try to reduce noise by only examining lattice paths coinciding with valid words (from a dictionary of 50K English words), the phonetically closest surviving path is “co|z|y mo the de|an”, still acoustically distant for rewriting. By contrast, another path, “co|z|im|ob|ina|ody|an”, produces a near-perfect phonetic match, provided that wordpieces are converted into phonemes directly, avoiding any word form. This motivates our approach.

\subsection{Learning Wordpiece-to-Phonemes Alignments}
In order to make phoneme predictions directly from wordpieces (independent of left and right history), we computed a mapping from wordpieces to associated phoneme sequences by learning their alignments on unigrams:

        \begin{enumerate}
            \item Start with 3.6 million words $w_i$ with frequency $f \geq 50$ from an online US English web corpus.
            \item For each $w_i$, retrieve pronunciations (phoneme sequences $p_{ij}$) from a phonetic dictionary.
            \item Segment each $w_i$ into a wordpiece sequence $wp_i$ using a wordpiece model (WPM).
            \item Use \{$wp_i$, $p_{ij}$\} frequency-weighted pairs to train IBM Translation Model 2~\cite{Brown_undated-gt}, using expectation maximization (EM) to learn alignments between wordpieces and phonemes.
            \item Retain the most frequent 10,000 wordpiece-to-phonemes alignments, with at least one mapping for each unique wordpiece emitted by the WPM in Step 3. \footnote{This produced mappings for 95.1\% of the total wordpiece vocabulary; the remaining 4.9\% of wordpieces were rarely observed in practice, but our algorithm could be extended to include them.}
        \end{enumerate}

\medskip
Table~\ref{tab:learned_alignments} gives examples of learned alignments. The mappings were compiled into a cyclic FST with  wordpieces on input labels and phonemes on output labels.  Composing a wordpiece lattice with this FST produces a phoneme lattice.

\begin{table}[h]
    \centering
    \begin{tabular}{c|l}
         \toprule
    Wordpiece  & Phoneme sequence(s) \\
    \midrule
    \pmboxdrawuni{2581}please & $\slash$p l i z$\slash$ \\
    \pmboxdrawuni{2581}for &  $\slash$f O r$\backslash$$\slash$, $\slash$f @\textasciigrave$\slash$, $\slash$f O$\slash$ \\
    ward & $\slash$w @\textasciigrave\ d$\slash$, $\slash$w O r$\backslash$ d$\slash$, $\slash$u @\textasciigrave\ d$\slash$ \\
    the & $\slash$T i$\slash$, $\slash$T$\slash$, $\slash$T E$\slash$, $\slash$D$\slash$, $\slash$T i$\slash$, $\slash$T @$\slash$, $\slash$t E$\slash$ \\
    e & $\slash$i$\slash$, $\sslash$, $\slash$eI$\slash$, $\slash$@$\slash$, $\slash$E$\slash$ \\
        \bottomrule
    \end{tabular}
    \caption{Learned wordpiece-phoneme alignments.  A wordpiece may be a part of a longer word; \pmboxdrawuni{2581} starts a new word.}
    \label{tab:learned_alignments}
\end{table}

\subsection{Applying Wordpiece-to-Phonemes Alignments}

We start with a CTC wordpiece lattice whose arc weights are negative log of $P(wp \mid a)$, where $wp$ is a given wordpiece and $a$ is the audio signal within a local CTC time window. We are interested in computing $P(p \mid a)$, where $p$ is a phoneme sequence. This can be expressed as:

\begin{equation}
P(p \mid a) = \sum_{\forall wp} P(p \mid wp, a) P(wp \mid a)
\end{equation}

Since our wordpiece-to-phonemes alignments are not aware of audio, we approximate $P(p \mid wp, a) \approx P(p \mid wp)$ and estimate the quantity $P(p \mid wp)$ as:

\begin{equation}
    P(p \mid wp) \approx 
\begin{cases}
    1,&      \text{if } \exists Align(p, wp)\\
    0,              & \text{otherwise},
\end{cases}
\end{equation}
\
\noindent where $Align$ represents our previously learned wordpiece-to-phonemes alignments. Our estimate is noisy and not locally normalized, but works in practice as we sum over many alignments and ignore alignments that do not lead to a match.

Let $W$ represent the wordpiece-to-phonemes FST constructed by the aforementioned procedure. Let $T'$ denote a grammar FST that identifies spans for rewriting in a wordpiece lattice.  $x'$ represents the wordpiece sublattice containing a potentially misrecognized phrase. The phoneme lattice is then computed as:

\begin{equation}
    x' \leftarrow x \circ_\sigma T'
\end{equation}
\begin{equation}
    z \leftarrow (x' \circ W) \circ_\sigma E,
\end{equation}
\noindent where $E$ is the expansion FST (Sec~\ref{sec:identifyingspans}). The resulting phoneme lattice $z$ can be further processed via steps described in Equations~\ref{eqn:phonemes_to_entity_fst} and \ref{eqn:recovery}.

\subsection{Log-Determinization}

When using phonemes to retrieve entities, one can consider the lowest-cost phonetic alignment leading to a match \cite{Serrino2019}, or consider all alignments. Considering all alignments yields better results possibly because individual alignments can have noisy scores due to bad paths in the non-autoregressive lattice and from applying G2P and phoneme substitution steps.

Using Equation~\ref{eqn:phonemes_to_entity_fst} produces an FST with alignments transducing from the source phonemes to the retrieved entities' graphemes.  We can sum over all alignments associated with a unique retrieved entity by projecting the alignment on output labels (containing entity graphemes), removing epsilons and determinizing the lattice in the log semiring:

\begin{equation}
h \leftarrow det_{LOG}(rmeps( \Downarrow(z \circ_\rho s')))
\label{eqn:recovery_log}
\end{equation}

\noindent This is similar to Equation~\ref{eqn:recovery} but now computes the sum:

\begin{equation}
P(w|a) = \sum_{ph} p(w|ph,a) p(ph|a)
\end{equation}

\noindent where $w$ is the grapheme sequence of a retrieved entity; $a$ is the audio signal within a local CTC time window; and each $ph$ corresponds to a phoneme sequence with a specific wordpiece-to-phoneme alignment.

FST determinization ensures that each accepted sequence is associated with a single path in the determinized FST, and its cost is the sum of the costs of all equivalent paths in the input lattice \cite{mohri2002weighted}.  Note the Sum operation differs between the tropical and log semiring: the tropical semiring takes the maximum of A and B, whereas the log semiring computes the true arithmetic sum (in the negative log space).  Determinization has a worst-case exponential runtime, but we avoid this due to a non-exponential number of unique paths in the determinized lattice (one per retrieved entity).

\begin{figure}[ht]
\centering
\includegraphics[width=\linewidth]{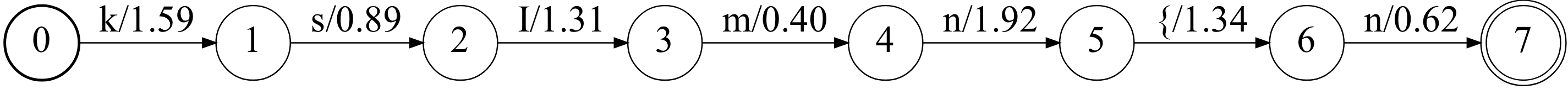}
\\[10pt]
\includegraphics[width=\linewidth]{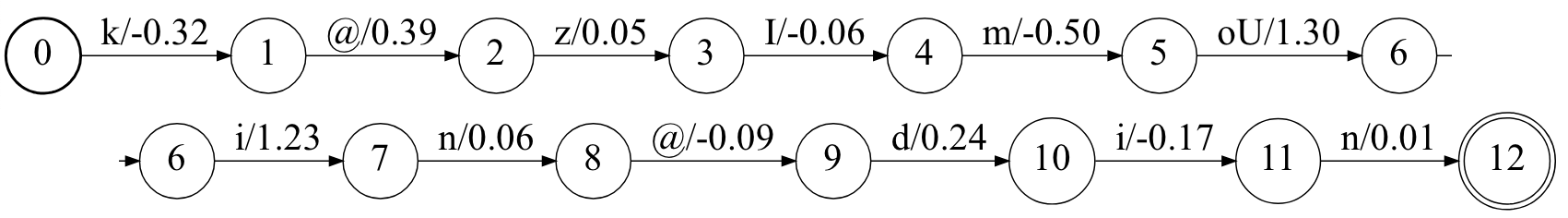}
\caption{Given a CTC lattice for a misrecognized utterance with the name "Kazi Mobi-Uddin", the top-panel shows the phonemes extracted without log-determinization (produced from the 1-best), while the bottom-panel shows the 1-best after transducing from wordpieces into phonemes and performing log-determinization of the entire phoneme lattice.  The correct phonemes are: /k A z i m oU b @ n u d @ n/.}
\label{fig:log_determinization_consensus}
\end{figure}

Below is an illustration of how log-determinization helps estimate phonetic consensus.  Given the CTC lattice for a misrecognition of "Kazi Mobin-Uddin" from Fig.~\ref{fig:comparison_of_lattices}, the baseline method (\cite{Serrino2019}) to estimate the phonetic signal is to select a grapheme path that matches a grammar and map its wordpieces to phonemes.  However, due to the noisy CTC signal this may produce a phoneme sequence quite distant from the truth (Fig.~\ref{fig:log_determinization_consensus}, top). However, if we instead transduce the entire wordpiece lattice into phonemes and then log-determinize the phoneme lattice (summing probability across all paths associated with the same phoneme sequence), the resulting 1-best phoneme path is very close to the true phonetics of "Kazi Mobin-Uddin" (Fig.~\ref{fig:log_determinization_consensus}, bottom).  In practice, it is much faster to log-determinize over the alignment sublattice (projected on graphemes) for matched entities only.

\subsection{Comparison Scoring}
For ASR spelling correction systems, scoring is needed to choose the best-fitting rewritten transcript, and also to avoid replacing an already correct ASR transcript with an incorrect rewrite.  For example, the decision to rewrite ``call Best Buy'' to ``call Beth Byer'' (personal contact name) should take into account which is a better fit acoustically.

To score the spell-corrected transcript against the ASR top transcript in the CTC lattice, we developed ``comparison scoring''; this is similar to 2nd-pass rescoring~\cite{NIPS2017_Xia}.  We treat the initial 1-best path as another retrieved entity, so its ``rewritten'' score comes from the log-determinized sum of all lattice paths which align phonetically to it.  Because retrieval works over a span captured by a rewriting grammar $T$, we likewise extract a subsequence of the 1-best path corresponding to $T$. As a result, a section of the 1-best path gets ``rewritten'' to itself, but with a new cost that reflects its acoustic fit to the entire CTC sublattice over the span tagged by $T$:

\begin{equation}
h'_0 \leftarrow inv(G) \circ h_0
\end{equation}
\begin{equation}
h"_0 \leftarrow det_{LOG}(rmeps( \Downarrow(z \circ_\rho h'_0)))
\label{eqn:recovery_comparison}
\end{equation}

\noindent where $h_0$ is a subsequence of the 1-best path in the ASR lattice tagged by grammar $T$; $h'_0$ is a phonemes-to-entity FST; and $h"_0$ is a path through the graphemes of $h_0$ whose cost is the sum of the negative log probabilities of all paths in the entire phoneme lattice which are consistent with the graphemes in $h_0$.

\section{EXPERIMENTAL SETUP}
\label{sec:experimental_setup}

We used an ASR system trained on proprietary training data consisting of 490K hours of anonymized short utterances, totaling 520 million English utterances with an average length of 3.4 seconds. Most of the utterances are machine-transcribed by a teacher ASR model; a small percentage are human-transcribed.

The input to the ASR system consists of 128-dimensional log Mel-filterbank energies, extracted from 32ms windows with 10ms shift. These are processed by two 2D-convolution layers, with strides $(2, 2)$, and the resulting features are fed to a stack of 32 Conformer layers~\cite{gulati2020conformer}. Each conformer has 8 attention heads with total dimension of 1536 and uses an intermediate FFN with a dimension of 384. The output CTC layer has a vocabulary of 4096 lowercase wordpieces plus the blank symbol.

For spelling correction, we tested two baseline approaches: 1) $n$-best extraction (based on unigram scores). Extract $n$-best paths, convert them to words, and perform phonetic entity retrieval from words. Higher values of $n$ show the advantage of using multiple paths, rather than just the best-scoring one. However, for dense CTC lattices, even $n$=1,000 only recalls a small fraction of paths. 2) Random path generation, where arcs were sampled by their unigram probability. This was meant to compensate for the tendency of $n$-best generation to return similar hypotheses to the 1-best, where only a few tokens were replaced by top alternatives.

Our experimental setup is direct wordpiece-to-phonemes transduction. This allows us to use the entire lattice depth while avoiding dependence on paths forming valid words.  We did not tune phonetic matching parameters.
\section{EVALUATION}
\label{sec:eval}

\begin{figure}[t]
\centering
\includegraphics[width=6.4cm]{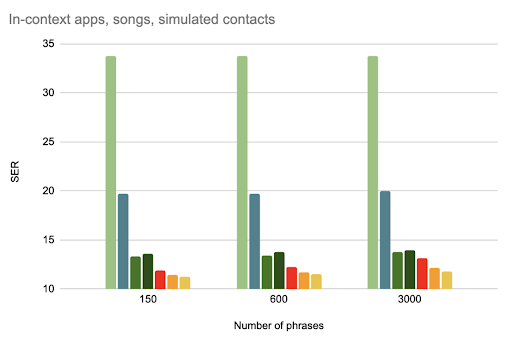}
\includegraphics[width=6.4cm]{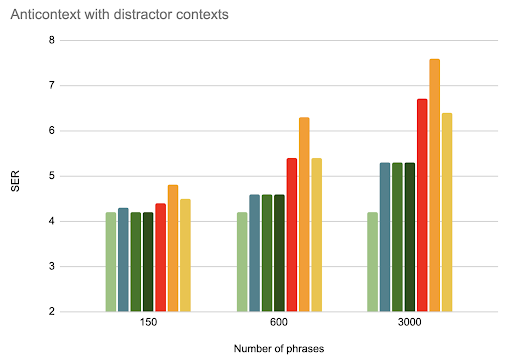}
\includegraphics[width=6.4cm]{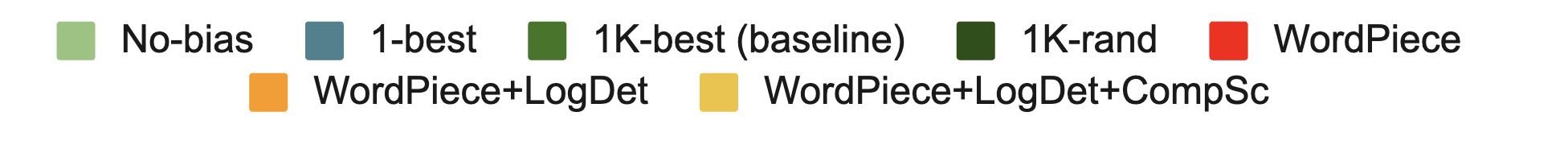}
\caption{In-context and anti-context SERs for lattice rewriting.  Baseline results are shown in green (we compare with 1K-best); experimental results are in red, orange, and yellow.}
\label{fig:wp_results}
\end{figure}
We evaluated our technique on an in-context dataset to measure biasing recall, and an out-of-context (or ``anti-context'') dataset to measure biasing precision. We follow recent literature in using Text-to-speech (TTS) for generating evaluation datasets \cite{peyser2019,wang2022b} to both safeguard user privacy and ensure alignment between the text and the audio. Our in-context dataset consists of 2,611 TTS utterances with entities from apps, songs, and simulated contacts (ACS), where the utterances' transcripts consist of the entity name following a type-appropriate prefix, like ``open'' for apps, ``play'' for songs, and ``call'' for contacts. Our anti-context set consists of 1,000 TTS anonymized utterances simulating general voice search queries.
Each test set is evaluated with a variable number of contextual entities from zero to 3,000, where the non-truth (``distractor'') entities are sampled using the same procedure as was used to create the entities for the in-context test set.

\subsection{$n$-best and $n$-random path baselines}
Our baseline replicates the word lattice rewriting technique in \cite{Serrino2019}, but here we apply it to a CTC wordpiece lattice by making the following adjustment. We extract $n$ paths for $n \in \{1, 1000\}$ from the wordpiece lattice. Each path is glued into words, which are assembled into a word lattice to be processed by the baseline spelling correction algorithm.  We found increasing $n$ to 1,000 greatly improves in-context SER due to higher recall of the phonetic ground truth. Still, this approach doesn't scale to the path count of dense CTC lattices.

 \begin{table}[t]
     \centering
     \footnotesize
     \begin{tabular}{l||l|l}
      \toprule
      \multicolumn{3}{c}{1-best (expt) vs. No-bias (base)} \\ \midrule
      W & \textbf{call annmaria de mars} & call Ann Maria Demars \\
      L & call Corey Allman & \textbf{call Corey Allmond} \\
     \bottomrule
     \multicolumn{3}{c}{1K-best (expt) vs. 1-best (base)} \\ \midrule
     W & \begin{tabular}{@{}l@{}} \textbf{can you play Agnus Dei} \\ \textbf{please} \end{tabular} & \begin{tabular}{@{}l@{}} can you play on you day \\ please \end{tabular} \\
     L & open MS marcadores & \textbf{open Mis marcadores} \\
     \bottomrule
     \multicolumn{3}{c}{WordPiece (expt) vs. 1K-best (base)} \\ \midrule
     W & \begin{tabular}{@{}l@{}} \textbf{can you play Flagpole} \\ \textbf{Sitta please} \end{tabular} & \begin{tabular}{@{}l@{}} can you play flagole sa \\ please \end{tabular} \\
     L & \begin{tabular}{@{}l@{}} can you play weird tour \\ please \end{tabular} & \begin{tabular}{@{}l@{}} \textbf{can you play} \\ \textbf{weirdcore please} \end{tabular} \\
     \bottomrule
     \multicolumn{3}{c}{WordPiece+LogDet (expt) vs. WordPiece (base)} \\ \midrule
     W & \begin{tabular}{@{}l@{}} \textbf{can you call Jason jixuan} \\ \textbf{hu please} \end{tabular} & \begin{tabular}{@{}l@{}} open can you call Jason G \\ schwenu please \end{tabular} \\
     L & call lirence P kelly & \textbf{call Lawrence P Kelly} \\ 
     \bottomrule
     \multicolumn{3}{c}{WordPiece+LogDet+CompSc (expt) vs. WordPiece+LogDet (base)} \\ \midrule
     W & \textbf{call Cornellius A. Tobias} & call carlius a tobias \\
     L & call Ave Chivas & \textbf{call Aven chievous} \\
     \bottomrule
     \end{tabular}
     \caption{Wins (W) and losses (L).}
     \label{tab:wins_and_losses}
\end{table}

\subsection{Wordpiece-to-phonemes Transduction}
Performance on the baselines and wordpiece-to-phonemes models WP (wordpiece), WP+LogDet (log-determinization), and WP+LogDet+CompSc (comparison scoring) is in Fig.~\ref{fig:wp_results}.

WP improves phonetic ground truth recall over the 1,000-best baseline by using all lattice paths for matching, lowering in-context SER. However, matching on more paths also increases the likelihood of rewriting anti-context utterances, leading to two kinds of regressions: phonetically inaccurate rewrites due to noisy paths or divergent wordpiece pronunciations ($the$$\rightarrow$ $\slash$T i$\slash$ in ``\underline{the}n''; $\slash$D$\slash$ in ``ba\underline{the}''); and phonetically accurate rewrites of waveforms with multiple plausible interpretations (``weirdcore''/``weird tour''). It's worth noting the latter type of contextual rewrites may be desirable for some inputs, despite the test counting them as regressions.

WP+LogDet maintains the same phonetic recall, but precision improves through better ranking of matched entities by considering phonetic evidence across all paths. A side effect is a higher combined score of the top-scoring candidate, increasing the likelihood of rewriting even anti-context utterances, counted as anti-context regressions. However, the phonetic accuracy of such rewrites is higher than with WP.

WP+LD+CompSc doesn't change the recall or ranking of matched entities. However, it is less eager to rewrite the original transcript due to requiring the rewrite to be phonetically closer to the lattice. This suppresses phonetically noisy rewrites, improving anti-context SER. In-context precision also slightly improves due to a side effect: when simultaneously, (1) the transcript is correct without rewriting, and (2) the ranking of rewrite candidates fails to select the correct entity, CompSc keeps the transcript from being rewritten.

Providing more phrases results in more matches, stressing entity ranking, with more rewriting of anti-context utterances.

\section{CONCLUSION}
\label{sec:conclusion}

We introduced a technique for performing ASR spelling correction on non-autoregressive lattices generated when decoding with CTC models. Such lattices may be noisy because they contain incoherent word sequences, which presents challenges for a word-based spelling correction approach.  
Our contribution lies in combining multiple approaches a) Using IBM Translation Model 2 to learn wordpiece-to-phoneme alignments; b) Proposing a method that can work on non-autoregressive lattices; c) Efficient scoring over exponentially many paths, which enables us to improve SER by approximating the acoustic ground truth using only graphemic outputs from the model. This approach is a simpler, modular and interpretable approach than prior spelling correction approaches that depend on features derived from acoustic inputs.

\bibliographystyle{IEEEbib}
\bibliography{mybib}

\end{document}